\documentclass[letterpaper, 10 pt, conference]{ieeeconf}  % Comment this line out if you need a4paper
\usepackage{graphicx}
\graphicspath{ {images/} }
\usepackage{amsmath}
\usepackage{xcolor}
\usepackage{floatrow}
\usepackage{subfig}
\usepackage{epstopdf}
\usepackage{hyperref}
\newtheorem{theorem}{Lemma}

\long\def\ocj#1{\textcolor{red}{#1}}

\long\def\ignore#1{ }

\IEEEoverridecommandlockouts                              % This command is only needed if 
\title{\LARGE \bf
Plenoptic Monte Carlo Object Localization \\ 
for Robot Grasping under Layered Translucency
}

\author{Zheming Zhou \hspace{0.5cm} Zhiqiang Sui \hspace{0.5cm} Odest Chadwicke Jenkins
\thanks{Z. Zhou, Z. Sui, and O.C. Jenkins are with the Department of Electrical Engineering and Computer Science, University of Michigan, Ann Arbor, MI, USA, 48109-2121
        {\tt\small [zhezhou|zsui|ocj]@umich.edu}}
}

\begin{document}

% \makeatletter
% \let\@oldmaketitle\@maketitle% Store \@maketitle
% \renewcommand{\@maketitle}{\@oldmaketitle% Update \@maketitle to insert...
%   \includegraphics[width=\linewidth]
%     {figures/teaser.png}\bigskip}% ... an image
% \makeatother

\maketitle
\thispagestyle{empty}
\pagestyle{empty}

%%%%%%%%%%%%%%%%%%%%%%%%%%%%%%%%%%%%%%%%%%%%%%%%%%%%%%%%%%%%%%%%%%%%%%%%%%%%%%%%
\begin{abstract} 
% \ocjf{I am not sure our results justify claims on multiple layers of translucency}
% \ocjf{revised abstract to be less of an introduction}
In order to fully function in human environments, robot perception needs to account for the uncertainty caused by translucent materials.  Translucency poses several open challenges in the form of transparent objects (e.g., drinking glasses), refractive media (e.g., water), and diffuse partial occlusions (e.g., objects behind stained glass panels).  This paper presents Plenoptic Monte Carlo Localization (PMCL) as a method for localizing object poses in the presence of translucency using plenoptic (light-field) observations.  We propose a new depth descriptor, the Depth Likelihood Volume (DLV), and its use within a Monte Carlo object localization algorithm.  We present results of localizing and manipulating objects with translucent materials and objects occluded by layers of translucency.  Our PMCL implementation uses observations from a Lytro first generation light field camera to allow a Michigan Progress Fetch robot to perform grasping.

%We propose of translucent objects with parti six\ocjm{spellout numbers less than 10} degree-of-freedom pose of such objects

%the robotic perception system is required to localize the objects with transparent surface and even the objects that behind those transparent surfaces. Under this circumstance, depth-sensor-based pose estimation methods will fail because of two reasons:1) RGB\_D sensors will fail to perceive the transparent surfaces and object behind; 2) a determinate depth map is not suitable for describing a scene with multiple transparent surfaces. To address this problem, we proposes a light-field-based perception framework to perform visible object 6DOF pose estimation in the scene with a new depth descriptor call depth likelihood volume(DLV). We deployed our approach on Fetch robot with commercial-level light field camera Lytro and established successful manipulation over transparent objects and the object behind window film covered glass. 
\end{abstract}

%%%%%%%%%%%%%%%%%%%%%%%%%%%%%%%%%%%%%%%%%%%%%%%%%%%%%%%%%%%%%%%%%%%%%%%%%%%%%%%%
\section{Introduction}
% Objects with translucent 

From frosted windows to plastic containers to refractive fluids, translucency is prevalent in human environments.  Translucent materials are commonplace in our daily lives and households, but remain an open challenge for autonomous mobile manipulators.  Previous methods, such as work by Foster et al.~\cite{foster2013visagge}, have enabled robots to navigate autonomously in the presence of glass and transparent surfaces.  When handling objects, however, robot perception systems must contend with a wider diversity of objects and materials.  

Translucent objects, in particular, break many of our assumptions in robot sensing and perception about opacity and transparency.  For example, existing six-DoF pose estimation methods~\cite{Suietal_SUM,narayanan2016perch} often heavily rely on RGB-D sensors to reconstruct 3D point clouds.  Such sensors are typically ill-equipped to handle the uncertainty caused by the reflection and refraction properties of translucent materials. As a result, translucent objects are often invisible to the robots for the purposes of dexterous manipulation.

%Domestic environment evolving a lot of transparent objects and most of those transparent objects take the function of containers like glass jar, glass cup, etc. In order to perform robot manipulation over those objects, a 6DoF pose estimation is required to yield not only the pose of transparent objects but also the object behind. 

\ignore{
For example, existing six-DoF pose estimation methods~ \cite{Suietal_SUM}~\cite{narayanan2016perch} often heavily rely on RGB-D sensors and some form of object matching against 3D point clouds.  
%using pointcloud to pointcloud matching method. 
However, in the scenes with transparent surface, RGB-D sensors will fail to give correct observations for those surfaced and the objects beyond which makes the pose estimation for those objects a very challenge problem.
}

An important topic related to this problem is multi-layer stereo depth estimation as studied by Borga and Knutsson~\cite{borga1999estimating}.  These findings establish that even transparent surfaces will emit their own distinguishable patterns.  When the pattern from translucent surfaces interacts with patterns from Lambertian surfaces,  the result will be multi-orientation epipolar image lines in multi-view stereo images. These stereo images can record these patterns within light fields, and equip a robot with the ability to identify surfaces with transparent properties.

Light field photography offers considerable potential for robot perception in scenes with translucency.  For example, Oberlin and Tellex~\cite{oberlintime} found that a high-resolution camera on the wrist of a robot manipulator can capture light fields for a static scene.  By moving the robot end-effector in a designed trajectory, this time lapse approach to light field capture was demonstrated as capable of manipulating transparent and reflective objects.  We now aim to extend similar ideas to a larger class of translucent materials, along with explicit pose estimation for more purposeful object manipulation.
%to generate light field photography. However, this process requires the target objects remains static to guarantee the light field information for target objects unchange over time. A nature way to overcome this is to deploy spatial multiplexing light field camera which can capture multiple  stereo images in a single shot. In our set up, we installed the first generation Lytro \cite{georgiev2013lytro} on the end effector of Fetch robot to perform light field capturing.

\begin{figure*}[!t]
\includegraphics[width=0.95\columnwidth]{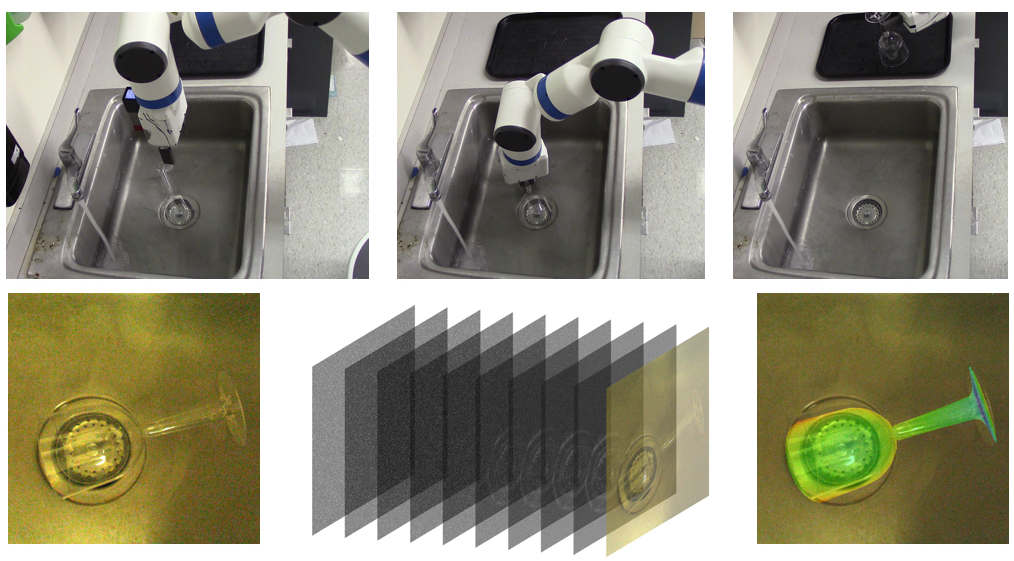}
\caption{(Top row) a robot equipped with a wrist-mounted light field camera correctly localizing, grasping, and placing a clear drinking glass from a sink of running water. (Bottom row) this grasp is performed by Plenoptic Monte Carlo Localization on the observed center view image (left), which computes a Depth Likelihood Volume (middle) to localize the object (right) through generative inference. 
%Middle and Right: A robot is picking up the glass cup using the estimate from PMCL and placing it onto the tray. (Bottom) Left: The center view image of the observations Middle: Depth Likelihood Volume calculated from the sub-aperture images. Right: Pose estimate from PMCL.  
}
\label{exp:sink}

\end{figure*}

In this paper, we propose Plenoptic Monte Carlo Localization (PMCL) as a method for six-DoF object pose estimation and manipulation under uncertainty due to translucency.  Our PMCL method uses observations from light field imagery collected by a Lytro camera mounted on the wrist of a mobile manipulator.  These observations are used to form a new plenoptic descriptor, the Depth Likelihood Volume (DLV).  The DLV is introduced to describe a scene with multiple layers of depth due to translucency.  The DLV is then used as a likelihood function with a Monte Carlo localization method for our PMCL algorithm to estimate object poses. 
%The contribution of this paper is: 1) A new depth descriptor for layered translucent environment 2) A Monte Carlo localization framework for object six-DoF pose estimation under layered translucent environment.
We demonstrate the efficacy of PMCL with DLV for manipulation in translucency with an implementation using a Michigan Progress Fetch robot.  We present results of object localization and grasping for two situations: transparent objects in transparent media (Figure~\ref{exp:sink}) and opaque objects diffusely occluded by translucent media.

\section{Related Work}
%\ocj{this entire section has been updated}
\subsection{Perception for Manipulation}
The problem of perception for manipulation remains challenging for robots working in human environments and the natural world.  The presented concepts for PMCL build on a substantial body of work in this area, which we summarize briefly.
%Robotic perception for manipulation in multi-translucent environment is a challenged problem and rarely addressed. We summarize a relevant subset of existing work in general scope respect to perception for manipulation. 
Ciocarlie et al.~\cite{ciocarlie2014towards} proposed a robust pick-and-place pipeline for the Willow Garage PR2 robot. This pipeline segments and clusters points which comprise isolated opaque tabletop objects observed from an RGB-D sensor. For more cluttered environments, Collet et al.~\cite{MOPED} proposed the MOPED perception framework for localizing objects by discriminatively clustering multi-view features in color images. Narayanan et al.~\cite{Narayanan-RSS-16} %~\cite{narayanan2016perch}~\cite{narayanan2017deliberative}
%\cite{narayanaswamy2011visual} 
take a deliberative approach to infer the pose of objects in clutter from RGB-D observations. This work performs A* search over possible scene states using a discriminative algorithm for 3D pose estimation. Similar in its aims, Sui et al.~\cite{Suietal2017ijrr,Suietal_SUM} have proposed generative models for scene inference and estimation.  Such generative models combine object detection from neural networks with Monte Carlo localization algorithms in the scenario of object sorting on highly cluttered tabletops.  

For transparent object perception, McHenry et al.~\cite{mchenry2006geodesic,mchenry2005finding} have used reflective features from transparent objects for segmentation in a single RGB image. Lei at al.~\cite{lei2011transparent} segment out transparent objects by searching failure detection from laser rangefinding (LIDAR) combined with RGB image features. Methods by Phillips et al.~\cite{phillips2016seeing} describe detection and estimation of rotationally symmetric transparent objects using edge features. 
Lysenkov et al.~\cite{lysenkov2013recognition} perform six-DoF pose estimation of transparent objects based on a silhoutte model corresponding with invalid RGB-D depth measurements.  Partial opacity from translucent materials can be problematic for such methods, where clear edge features become blurred due to diffuse reflection.
%But their pose estimation pipeline cannot deal with a scene when objects are blocked by translucent surfaces and they have strong assumption of object standing on the known support plane.

\begin{figure*}[thpb]
\vspace{+1em}
   \centering
      \includegraphics[width=0.95\textwidth]{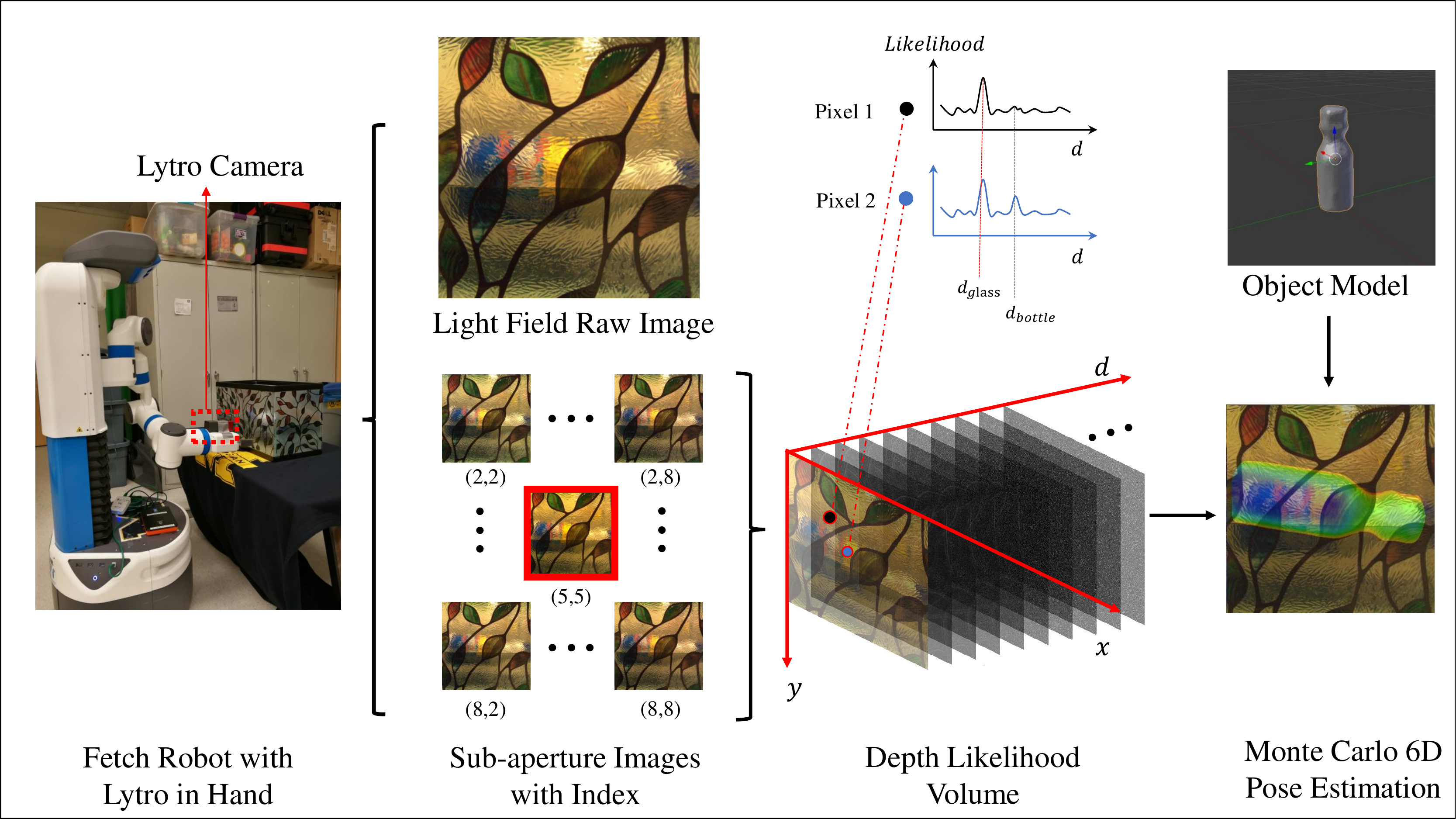}
      \caption{ An overview of our Plenoptic Monte Carlo Localization framework. A light field camera is installed on the end effector of the robot. After taking a single shot light field image of the scene, sub-aperture images are extracted (center view highlighted in red). The depth likelihood volume (DLV) is then computed as a 3D array of depth likelihoods over certain pixels $(i,j)$ for depth $d$.  The DLV is a comparator of color and gradient similarity between the center view and other sub-aperture images. Assuming a known geometry and region of interest, the six-DoF object pose is estimated by Monte Carlo Localization over a constructed DLV.}
      \label{overview}

\end{figure*}

\subsection{Light Field Photography}
The contributions of this paper are founded upon models described by Levoy and Hanrahan~\cite{levoy1996light} for understanding light fields and plenoptic functions.  Their seminal paper covers the foundations of capturing light fields from digital imagery and using them to synthesize new viewpoints from arbitary camera positions. Building on this work, microlens-based light field photography~\cite{georgiev2013lytro,ng2006digital} has witnessed significant advancements in depth estimation, image refocusing, transparent object recognition, and surface reconstruction.

In computer vision, Maneo et al.~\cite{maeno2013light} proposed 
%\ocjf{quotations in latex use ``directed double single quotes''}
``light field distortion features'' to capture distortions and recognize transparent objects. Sulc et al.~\cite{sulc2016reflection} separates diffuse color components from 4D light field imagery to suppress non-lambertian surface's reflection.  Wang et al.~\cite{wang2015occlusion} introduced a light field occlusion model for accurate recovery of the depth information around the edge where occlusion occurs.  Jeon et al.~\cite{jeon2015accurate} overcome the narrow baseline problem of light field cameras based on the sub-pixel shift method. This method generates accurate depth images even when the displacement of two adjacent sub-aperture images is less than 1 pixel. Our presented methods for PMCL build directly upon ideas in recent work by Goldluecke et al.~\cite{reconst_accv2017,wanner2013reconstructing} for 3D reconstruction in multi-translucent environments.  This work proposes generating multi-orientation features observed in epipolar plane images generated by a light field imagery, with impressive results for 3D reconstruction in high translucency.

In robotics, Oberlin and Tellex~\cite{oberlintime} introduced a time lapse approach to capture light for pick-and-place localization with a Rethink Baxter robot. This work demonstrated compelling results for localizing grasp and placement points in scenes with transparency and reflection, which has been problematic for current sensors.
%\ocj{minor details} Both their work and ours uses light field images to localize objects in the scene. However, their primary goal is to use robot as agent to perform light field capturing and scene modeling. 
%Furthermore, their object localization process requires the target objects remains static to guarantee the light field information for target objects unchange over time. \ocj{secondary point}
Our PMCL method shares similar aims with more general models of translucency in mind.  Further, estimation of six-DoF object pose estimation by PMCL will allow for greater flexibility in planning and executing manipulation actions.  We posit PMCL to be readily capable of object tracking from plenoptic observations, although such experiments are left for future work.
\section{Problem Formulation}
% \ocj{express within tracking formulation for PMCL}
Given an input light field image observation $Z$, the purpose of six-DoF pose estimation is to infer the rigid transformation from an object's local coordinate frame $\mathcal{O}$ to the camera's coordinate frame $\mathcal{C}$. We assume as given the geometry of the target object $o$.  Formally, we aim to find the maximum likelihood estimate for the object's pose $q$ given $o$ and a map representation $m$ in 3D world coordinates:
\begin{equation}
\label{fomulation}
\arg\max_q P(q|m,o)
\end{equation}
The map $m$ is often computed as a metric representation, such as a 3D reconstruction or point cloud.  In the case of common RGB-D cameras, the map representation is a one-to-one mapping from locations in 3D space $(x,y,z)$ into depth value $d$ at pixel index $(i,j)$ of a depth image.  Such a one-to-one mapping assumes opacity in that the sensed depth at a particular pixel is due to light from only one object.

We propose the {\bf Depth Likelihood Volume (DLV)} as an alternative one-to-many mapping to consider the likelihood of a pixel over multiple levels of depth.  As the case for translucent objects, the DLV representation is advantageous in environments where multiple objects at more than one depth are responsible for the light sensed at a pixel.
%different from using determinate depth map(image) to represent the depth information of a scene, we argue that a depth likelihood volume(DLV) which describe how likely a depth in 3D workspace space corresponding to certain location in the 2D image plane is more suitable for scene representation when transparent surfaces occurs. 
The DLV representation expresses $m$ as the mapping:
\begin{equation}
\label{dlm_eq}
m:\mathcal{M}_\rho(x,y,z) \rightarrow L(i,j,d)
\end{equation}
where $\mathcal{M}_\rho(x,y,z)$ represents a 3D point $(x,y,z)$ along a light ray $\rho$ taken as input.  The output $L(i,j,d)$ is the likelihood of light along the ray $\rho$ emitted from depth $d$ being received by pixel $(i,j)$ in the image plane. 
%$(i,j)$ is the pixel index that the light ray $\rho$ hits on the image plane. 
For our light field cameras, we assume the image plane is determined by the center view image of the sub-aperture images extracted from light field observation $Z$. $d$ is discretized possible depths along light ray $\rho$. An overview of our approach to this problem is shown in Figure~ \ref{overview}.

\ignore{
Further introduction of DLV and its corresponding property is explained in Section \ref{DLV}. The method for building DLV using light field image is proposed in  Section \ref{building_DLV}. The Monte Carlo method for inferring object's 6D pose is in Section \ref{PPF}.
}

\begin{figure*}[!t]
\vspace{1em}
   \centering
      \includegraphics[width=0.620\textwidth]{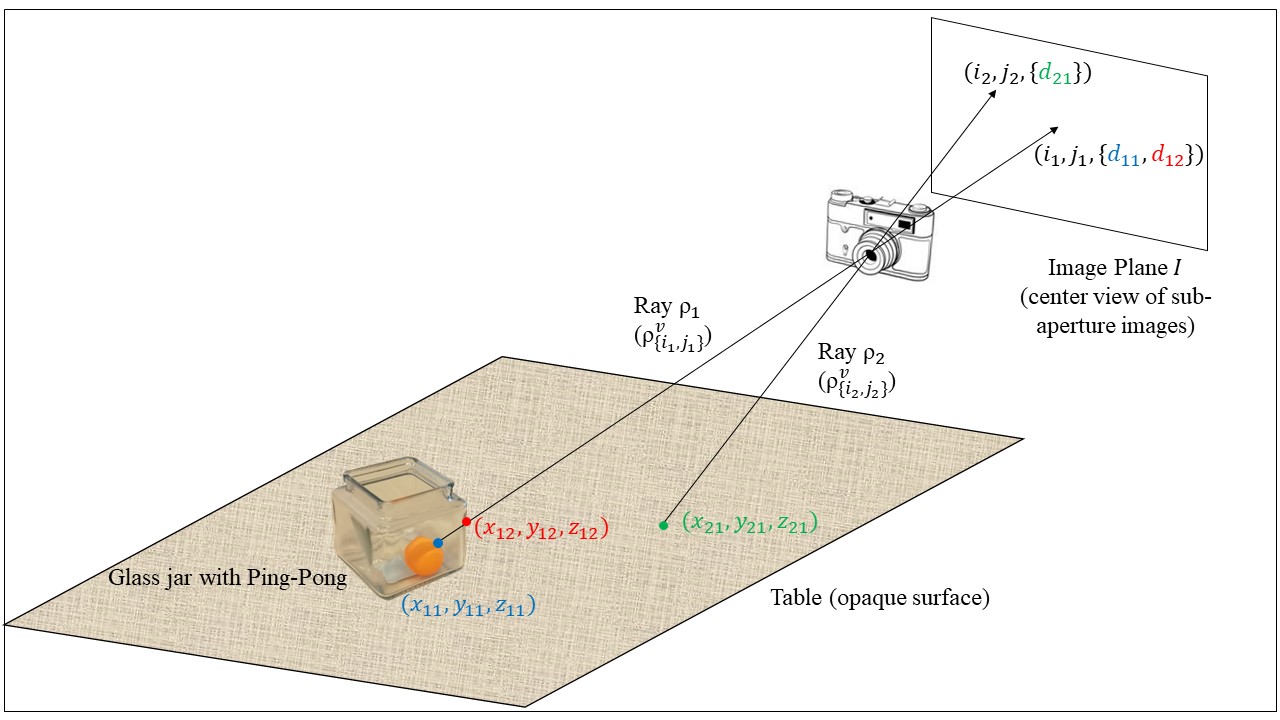}
      \includegraphics[width=0.35\textwidth]{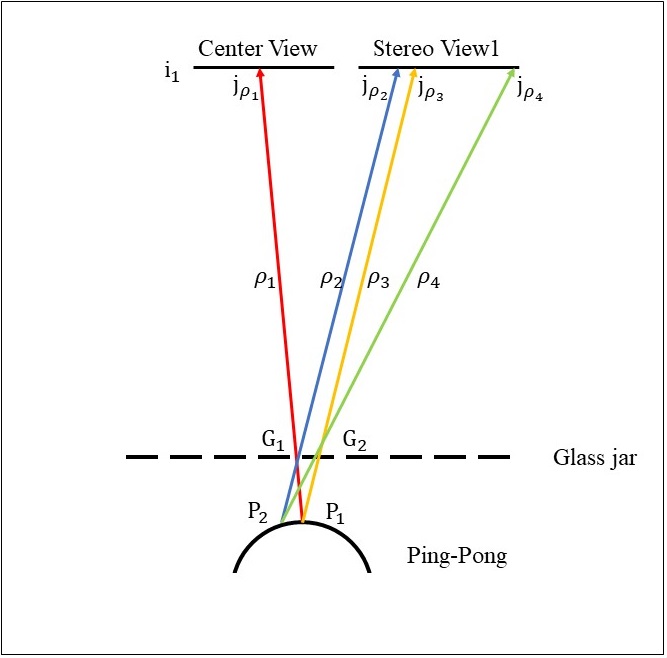}
      \caption{(Left) a scene with a transparent glass jar containing a ping-pong ball at rest on an opaque table. Along ray $\rho_1$, two surfaces (incident to the ball and the front surface of the jar) contributes to the pixel value, while along ray $\rho_2$ only one surface (incident to the table) appears.  (Right) a planar top-down view of rays incident to the ball and the jar. The center view image plane, $(i_1,j_{\rho_1})$ receives a weighted sum of light rays reflected from both the glass surface point $G_1$ and the ping-pong surface $P_1$.  Three example rays 
      %\ocj{(are these three $\rho_{2,3,4}$ different examples to consider?)} In the stereo view 1, three rays 
corresponding to $\rho_2$ (reflection of the surface from the glass jar), $\rho_3$ (reflection of the ping-pong ball through the glass), and $\rho_4$ (random ray) received by the image plane with incidence to scene points $(G_1,P_2)$, $(G_2,P_1)$, and $(G_2,P_2)$, respectively. They indicate three depth $d_g,d_p,d_i$ when form stereo pair with ray $\rho_1$.}
      \label{depth_dist}
   \end{figure*}

\ignore{
\begin{figure}[!t]
   \centering
      \includegraphics[scale=0.4]{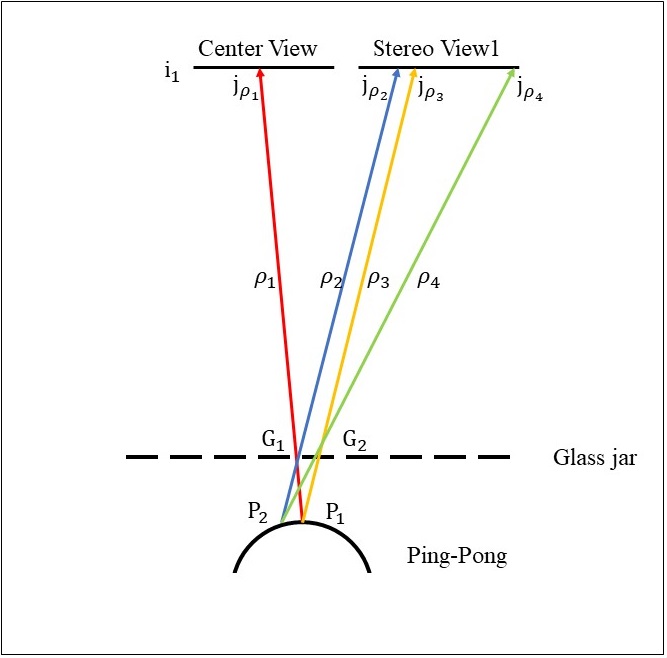}
      \caption{\ocj{try to keep this and previous figure at the top of the same page.} A 2D front view of ping-long and glass jar environment setting. In center view plane, $(i_1,j_{\rho_1})$ receives light rays from glass surface point $G_1$ and ping-pong surface $P_1$. In the stereo view 1, three rays $\rho_2,\rho_3,\rho_4$  received by image plane have ray components $(G_1,P_2)$, $(G_2,P_1)$, and $(G_2,P_2)$ respectively. They indicate three depth $d_g,d_p,d_i$ when form stereo pair with ray $\rho_1$ in the center view. }
      \label{2D_view}
   \end{figure}
}
 
 \section{Depth Likelihood Volumes}

Before presenting our PMCL method for pose estimation, we first define the Depth Likelihood Volume.  We describe the properties of the DLV for distinguishing multiple depths at a given point in an image due to translucency.  The construction of the DLV and its use for pose localization is described in the following section.

%In this section, we present the definition and property of DLV and prove its ability to distinguish multiple depths in a scene. We show how to construct DLV using light field sub-aperture images and the particle optimization algorithm building on DLV for 6D pose estimation. 

\subsection{Formulation}
\label{DLV}
Given a known 3D workspace and its corresponding center view sub-aperture image plane $I$, a Depth Likelihood Volume is defined in Equation~\ref{dlm_eq}. The DLV makes the following basic assumptions and notations for the scene:
\begin{enumerate}
\item[(1)] Each surface point emits light rays $\rho$ in each channel as a Gaussian over $(r,g,b)$ with mean $(\mu_r,\mu_g,\mu_b)$ and variance $(\sigma_r^2, \sigma_g^2, \sigma_b^2)$ which means $\rho = \mathcal{N}(\lambda;\mu_c,\sigma_c^2), c\in\{r,g,b\}$, as similarly assumed by Oberlin and Tellex~\cite{oberlintime}. Under constant lighting condition we assume every point in the scene shares the same variance for the same color channel which means $\sigma_c = \sigma_c', c\in\{r,g,b\}$ for all points in the scene.
% $ (\sigma_{G1c}^2=\sigma_{G2c}^{2}=\sigma_{P1c}^2=\sigma_{P2c}^{2}=\sigma_c^2 ), c\in\{r,g,b\}$. 
\item[(2)] An observed bundle of rays located at pixel plane $(i,j)$ is a linear combination of all light rays emitted by surface points along the light rays with the normalization scalers $\alpha_i$. $\alpha_i$ indicates the percentage of rays emitted by the surface in observed rays which measures the transparency of the surface, and we have $\sum_i \alpha_i = 1$.
% \begin{equation}
% \label{alpha_sum}
% \sum_i \alpha_i = 1
% \end{equation}
\end{enumerate}

Consider the example in Figure \ref{depth_dist} (Left) of two light rays $\rho_{\{i_1,j_1\}}^{v},\rho_{\{i_2,j_2\}}^{v}$ imaged by the central view sub-aperture image.  The index $v$ indicates center view, and $\{\cdot,\cdot\}$ are pixel coordinates in the center view. These rays are in the 3D space hitting the center view plane $I$ at $(i_1,j_1),(i_2,j_2)$, respectively.  Along $\rho_{\{i_1,j_1\}}^{v}$, there are two surfaces emitting light  which are sensed by the central view: one is a ping-pong ball and the other is the glass jar.  In contrast, along $\rho_{\{i_2,j_2\}}^{v}$, only light emitted by the table is sensed in the central view. Then $\rho_{\{i_1,j_1\}}^{v},\rho_{\{i_2,j_2\}}^{v}$ can be expressed respectively as:
\begin{equation}
\begin{aligned}
&\rho_{\{i_1,j_1\}}^{v} = \alpha_g\rho_{\textrm{glass}} + \alpha_p\rho_{\textrm{ping-pong}} \\
&\rho_{\{i_2,j_2\}}^{v} = \alpha_t\rho_{\textrm{table}}
\end{aligned}
\end{equation}
where $\rho_{\textrm{glass}},\rho_{\textrm{ping-pong}},\rho_{\textrm{table}}$ represents the light rays emitted by glass, ping-pong, and table surfaces, respectively. According to our second assumption, we also have $\alpha_g + \alpha_p = 1$ and $\alpha_t = 1$.

% \begin{equation}
% \sum_{d\in\rho}  L(i,j,d) = 1
% \end{equation}

% As shown in Fig\ref{depth_dist}, a glass jar with ping-pong is put on a opaque table. For pixel $(i_1,j_1)$, two possible depths appears: one is generated by the surface of the glass jar, the other comes from ping-pong's surface. For pixel $(i_2,j_2)$, only one depth is valid which comes from the table surface. Then an ideal DLV will have the property for Figure \ref{depth_dist} (take $(i_1,j_1)$ as example) that:
% \begin{equation}
% \begin{aligned}
% &L(i=i_1,j=j_1,d=d_1)>0\\
% &L(i=i_1,j=j_1,d=d_2)>0\\
% &L(i=i_1,j=j_1,d=d_2)+L(i=i_1,j=j_1,d=d_1)=1\\
% &\forall d\neq d_1,d_2 \; L(i=i_1,j=j_1,d)=0\\
% \end{aligned}
% \end{equation}

% To construct a DLV, sub-aperture stereo images are required to generate depth and corresponding depth likelihood. Take Figure \ref{depth_dist} scenario as an example, after taking a light field image and extracted sub-aperture images, we can form a stereo pair using two sup-aperture images in the same row. Take a scanline $j_i$, we have 2D set up as shown in Figure \ref{2D_view}.     
Then the depth likelihood is defined as:
\begin{equation}
\begin{aligned}
\label{depth_likelihoos_function}
&L(i,j,d) =\\ 
&\sum_n\frac{\max_k ||\rho_{\{i,j\}}^{v} ,\mathcal{T}_k^n(\rho_{\{i,j\}}^{v})||^2 - ||\rho_{\{i,j\}}^{v} , \mathcal{T}_d^n(\rho_{\{i,j\}}^{v})||^2}{\sum_k ||\rho_{\{i,j\}}^{v} ,\mathcal{T}_k^n(\rho_{\{i,j\}}^{v})||^2
}\\
\end{aligned}
\end{equation}
where $\mathcal{T}_k^n(\rho_{\{i,j\}}^{v})$ is the transformation function finding the light ray corresponding to $\rho_{\{i,j\}}^{v}$ in stereo pair image index with $n$ that indicates depth $k$ . For light field camera known baseline $b$ and focal length $f$, the $\mathcal{T}_k^n(\cdot)$ can be expressed as $\frac{bf}{D}$, where $D$ is disparity which is the function of $n$ and $k$. $||\cdot,\cdot||^2$ is the squared similarity distance between two light rays over $\{r,g,b\}$ color space which is defined as $L_2$ distance between two Gaussian mixture models according to assumption (1) and (2) and can be expressed as Equation~\ref{similarity_function}. 

\subsection{Validity}

We claim that for a given $(i,j)$ in DLV the following Lemma holds:
\begin{theorem}
$$\alpha_1 < \alpha_2 \iff L(i,j,d_{1}) < L(i,j,d_{2})$$

\end{theorem}
where $d_{1},d_{2}$ indicates the true surface depth viewed from center view with transparency indicator $\alpha_1,\alpha_2$. This means, the more transparent a surface, the less likelihood the depth of this surface will be in the DLV.
% where $d_{nt}$ is the true depth of opaque surface, $d_{t}$ is the true depth of transparent surfaces, $d_{others}$ is the other false depth. Take Figure \ref{2D_view} as example, for $(i_i.j_1,d)$ in DLV, $d_{nt} = \mathcal{D}(j_{\rho_1},j_{\rho_3})$, $d_{t} = \mathcal{D}(j_{\rho_1},j_{\rho_2})$, and $\mathcal{D}j_{\rho_1},j_{\rho_4 })\in d_{others}$. Here $\mathcal{D}(\cdot)$ means function map disparity to depth which can be expressed as $\frac{bf}{D}$, where $b$ is baseline of two stereo images, $f$ is focal length, and $D$ is disparity. 
% To illustrate this, we first give proves than follow with intuitive explanations.

To show the Lemma 1, we consider the scene as shown in Figure \ref{depth_dist} (Right). In the center view (where DLV will be built), $\rho_{\{i,j_{\rho_1}\}}^{v}$ (simplify notation as $\rho_1$) contains rays from the glass surface point $G_1$ and ping-pong surface point $P_1$ which has depths $d_g$, $d_p$ respectively. We then evaluate three possible depths in this scene: $d_g$, $d_p$, and a invalid depth $d_i$. For every surface point, corresponding $\alpha_g,\alpha_p,\alpha_i$ are set as  $\alpha_g = \alpha,\alpha_p = （1-\alpha）,\alpha_i = 0$. Notice that $\alpha < 0.5$ since glass is a transparent surface while ping-pong is not. Using function $\mathcal{T}_k^n(\rho_1)$ we can find three rays ($\rho_2$,$\rho_3$,$\rho_4$) in stereo image $n$ corresponding to three depths $d_g$, $d_p$, and $d_i$ separately. Then, we can write ray $\rho_1$ as:

\begin{equation}
\rho_1  = \alpha \mathcal{N}(\lambda;\mu_{G_1c},\sigma_{G_1c}^2) + (1-\alpha)\mathcal{N}(\lambda;\mu_{P_1c},\sigma_{P_1c}^2)
\end{equation}
where $c\in\{r,g,b\}$ represents three color channels. Without loss of generality, we investigate the red channel and write $\rho_2,\rho_3,\rho_4$ in same fashion:
\begin{equation}
\rho_2  = \alpha \mathcal{N}(\lambda;\mu_{G_1r},\sigma_{G_1r}^2) + (1-\alpha)\mathcal{N}(\lambda;\mu_{P_2r},\sigma_{P_2r}^2)
\end{equation}
\begin{equation}
\rho_3  = \alpha \mathcal{N}(\lambda;\mu_{G_2r},\sigma_{G_2r}^2) + (1-\alpha)\mathcal{N}(\lambda;\mu_{P_1r},\sigma_{P_1r}^2)
\end{equation}
\begin{equation}
\rho_4  = \alpha \mathcal{N}(\lambda;\mu_{G_2r},\sigma_{G_2r}^2) + (1-\alpha)\mathcal{N}(\lambda;\mu_{P_2r},\sigma_{P_2r}^2)
\end{equation}

Here, we assume that transparent surfaces emit an equal amount of light rays between any two stereo images because the disparity range between adjacent sub-aperture views of the Lytro camera is smaller than $\pm 1$ pixel \cite{yu2013line} (around $10^{-4}$ rads in view angle in our experiment setting). The squared similarity ( $||\cdot , \cdot||^2$ ) distance between $\rho_1$ and any other rays can be expressed as:
\begin{equation}
\label{similarity_function}
||\rho_1(\lambda),\rho_n(\lambda)||^2 = \int(\rho_1(\lambda)-\rho_n(\lambda))^2   \;d\lambda
\end{equation}

where $n\in\{2,3,4\}$. Given this general expression of distance, we can now provide explicit expressions for the example shown in Figure~\ref{depth_dist} Right:
\begin{equation}
\label{Eq.rho1}
||\rho_1(\lambda),\rho_2(\lambda)||^2 =2(1-\alpha)^2(A-\mathcal{N}(\mu_{P_1r};\mu_{P_2r},2\sigma_r^2)) 
\end{equation}
\begin{equation}
\label{Eq.rho2}
||\rho_1(\lambda),\rho_3(\lambda)||^2 = 2\alpha^2(A-\mathcal{N}(\mu_{G_1r};\mu_{G_2r},2\sigma_r^2)) 
\end{equation}
\begin{equation}
\label{Eq.rho3}
\begin{aligned}
||\rho_1(\lambda),\rho_4(\lambda)||^2 &= ||\rho_1(\lambda),\rho_2(\lambda)||^2 + ||\rho_1(\lambda),\rho_3(\lambda)||^2 \\
& + 2\alpha(1-\alpha)(\mathcal{N}(\mu_{G_1r};\mu_{P_1r},2\sigma_r^2)\\
& - \mathcal{N}(\mu_{G_1r};\mu_{P_2r},2\sigma_r^2)\\
& + \mathcal{N}(\mu_{G_2r};\mu_{P_2r},2\sigma_r^2)\\
& - \mathcal{N}(\mu_{G_2r};\mu_{P_1r},2\sigma_r^2))\\
\end{aligned}
\end{equation}

where $A= \frac{1}{\sqrt{4\pi \sigma_r^2}}$ and given the following relation:
\begin{equation}
\int \mathcal{N}(x;\mu,\Sigma)\mathcal{N}(x;\mu',\Sigma')dx = \mathcal{N}(\mu;\mu',\Sigma + \Sigma')
\end{equation}

For the same object, under $~10^{-4}$ rads view difference, we assume the color difference between two surface points have the same scale $\Delta$. This assumption implies, for some small value $\epsilon$, that $ \epsilon > \Delta = |\mu_{P1r} - \mu_{P2r}| = |\mu_{G1r} - \mu_{G2r}|$.

Disregarding constant scale $2$, Equation~\ref{Eq.rho1}, \ref{Eq.rho2}, \ref{Eq.rho3} can be simplified as Equation~\ref{Eq.rho1_sim}, \ref{Eq.rho2_sim}, \ref{Eq.rho3_sim}:
\begin{equation}
\label{Eq.rho1_sim}
% \begin{aligned}
||\rho_1(\lambda),\rho_2(\lambda)||^2 = (1-\alpha)^2A(1-\exp^{-\frac{\Delta^2}{4\sigma_r^2}})
\end{equation}
\begin{equation}
\label{Eq.rho2_sim}
||\rho_1(\lambda),\rho_3(\lambda)||^2 = \alpha^2A(1-\exp^{-\frac{\Delta^2}{4\sigma_r^2}})
\end{equation}
\begin{equation}
\label{Eq.rho3_sim}
||\rho_1(\lambda),\rho_4(\lambda)||^2 = ((1-\alpha)^2 + \alpha^2)A(1-\exp^{-\frac{\Delta^2}{4\sigma_r^2}})
% \end{aligned}
\end{equation}
% where $B = \exp^{-\frac{1}{2\sigma_c^2}}$

% Notice for transparent surface, $\alpha < 0.5$ (approximate $0.2$ for normal glass surface), we have:
% \begin{equation}
% L_2(\rho_1(\lambda),\rho_4(\lambda)) > L_2(\rho_1(\lambda),\rho_2(\lambda)) > L_2(\rho_1(\lambda),\rho_3(\lambda))
% \end{equation}

% Then apply Equation~\ref{depth_likelihoos_function} (take $n=1$ since only one stereo pair in this example), we have

Considering an individual stereo pair and applying Equation~\ref{depth_likelihoos_function}, we can now express the DLV values for the possible depths for the surface of ping-pong ball, $d_p$, the glass surface, $d_g$, and the invalid depth, $d_i$, as:
\begin{equation}
L(i,j,d_p) = \frac{(1-\alpha)^2}{(1-\alpha)^2+\alpha^2}
\end{equation}
\begin{equation}
L(i,j,d_g) = \frac{\alpha^2}{(1-\alpha)^2+\alpha^2}
\end{equation}
\begin{equation}
L(i,j,d_{i}) = \quad \quad 0
\end{equation}
which implies that the ping-pong surface must return more light than the glass surface:
\begin{equation}
\alpha_g < \alpha_p \iff L(i,j,d_{g}) < L(i,j,d_{p}), \alpha_p,\alpha_g \in[0,1]
\end{equation}
Therefore, Lemma 1 holds. 

% The other way to view $Lemma 1$ is the appearance of multiple orientations of epipolar line in a epipolar plane image. \cite{2_ori} shows an example that with a reflective surface, a particular epipolar line that passes thought the same pixel has multiple possible orientations which corresponding to multiple depths in the image.
% \begin{figure}[thpb]
%    \centering
%       \includegraphics[scale=0.4]{EPI}
%       \caption{A scene with reflective surface and corresponding epipolar plane image for red scanline in center view}
%       \label{depth_dist}
%    \end{figure}

\subsection{Computation}
\label{building_DLV}
Our implementation uses the $L_2$ distance between adjacent pixel colors to approximate the similarity of rays in stereo pairs, as photosensors are unable to capture the distribution over wavelengths of light.
%, instead we use $L_2$ distance between two pixel colors assisted with the color gradient to calculate the similarity of rays in stereo pairs.
Considering this limitation, a cost-volume stereo comparison method based on sub-pixel shift \cite{hosni2013fast,jeon2015accurate} was implemented. Two different cost volumes were implemented: the sum of $L_2$ distance in color space ($C_c$) and the sum of gradient differences ($C_g$). The cost volume $C$ then can be defined as:
\begin{equation}
C(\mathbf{x}_\rho,l) = \beta C_{c}(\mathbf{x}_\rho,l) + (1 - \beta)C_{g}(\mathbf{x}_\rho,l)
\end{equation}
where $\mathbf{x}_\rho=  (i,j)$ describes the image coordinate of ray $\rho$, $l$ is depth labels and $\beta$ is a scalar to weight two parts. The terms $C_{c}$ and $C_{g}$ are defined as:
\begin{equation}
\begin{aligned}
C_{}&(\mathbf{x}_\rho,l)  = \\
& \sum_{\mathbf{s}\neq \mathbf{s}_c}\sum_{\mathbf{x}_\rho\in R_\mathbf{x}}\min(|I(\mathbf{s}_c,\mathbf{x}_\rho)- I(\mathbf{s},\mathbf{x}_\rho+\Delta \mathbf{x}(\mathbf{s},l))|,\tau_1)\\
C_{g}&(\mathbf{x}_\rho,l) =\\
&\sum_{\mathbf{s}\neq \mathbf{s}_c}\sum_{\mathbf{x}_\rho\in R_\mathbf{x}}\gamma\min(|I_x(\mathbf{s}_c,\mathbf{x}_\rho)- I_x(\mathbf{s},\mathbf{x}_\rho+\Delta \mathbf{x}(\mathbf{s},l))|,\tau_2)\\
  &+(1-\gamma)\min(|I_y(\mathbf{s}_c,\mathbf{x}_\rho)- I_y(\mathbf{s},\mathbf{x}_\rho+\Delta \mathbf{x}(\mathbf{s},l))|,\tau_2)
\end{aligned}
\end{equation}

where $I$ is the image, $I_x,I_y$ is the image gradient in $x,y$ direction, $R_\mathbf{x}$ is a rectangular region that center at $\mathbf{x}_\rho$; $\tau_1,\tau_2$ is a truncation value of a robust function, $\Delta\mathbf{x}(s,l)$ is the sub-pixel displacement, and $\gamma = \frac{|\mathbf{s}-\mathbf{s}_c|}{|\mathbf{s}-\mathbf{s}_c|+|\mathbf{t}-\mathbf{t}_c|}$ weights different sub-aperture's gradient contributions to the center view image. Variables $\mathbf{s} , \mathbf{t}$ represent pixel in sub-aperture image index coordinate and $\mathbf{s}_c , \mathbf{t}_c$ represent pixel in the center view.

For a certain depth label $l_i$, the depth likelihood can be expressed as below based on Equation~\ref{depth_likelihoos_function}:
\begin{equation}
L(\mathbf{x}_\rho,l_i) = \log(\frac{\arg\max_{l} C(\mathbf{x}_\rho,l) - C(\mathbf{x}_\rho,l_i)}{\sum_{l_i}( C(\mathbf{x}_\rho,l_i))} + 1)
\end{equation}
Optionally, to further distinguish possible depths, the DLV can be truncated by finding $N_{lm}$ number of local maximum with its $K_{lm}$ number of neighbors and setting the other depth likelihoods to $0$.
% \subsection{Particle Filtering over DLV}
% \label{PPF}
% Particle filtering is employed with each object estimator to infer the pose of transparent object  $q$. A particle filter is a means of inference for the sequential Bayesian filter in Eq.\ref{eq_pf} throught an approximation consisting of $n$ weighted particles, $\{q_t^{(j)},w_t^{(j)}\}_{j=1}$. Weight $w_t^{(j)}$ for particle $q_t^{(j)}$ is expressed as 
% \begin{equation}
% \label{eq_pf}
% Bel(q_t) \propto p(z_t|q_t)\int_{q_{t-1}}p(q_t|q_{t-1})Bel(q_{t-1})dq_{t-1}
% \end{equation}
% \begin{equation}
% Bel(q_t) \propto p(z_t|q_t)\sum_{j}p(q_t^{(j)}|q_{t-1}^{(j)})Bel(q_{t-1}^{(j)})
% \end{equation}
% as described by Dellaert et al. \cite{pf} The initial belief of object pose is uniform. At each time instance, the weight
% of each hypothesis is computed, normalized to one, and resampled based on importance into an updated set of $n$ particles:

\section{Plenoptic Monte Carlo Object Localization}
\label{PPF}
Building on the DLV, we now describe our method of object pose estimation as   Plenoptic Monte Carlo Localization.
PMCL employs particle filtering to estimate the pose of target objects from the computed DLV. PMCL takes direct inspiration from the work of Dellaert et al.~\cite{Dellaert_MCL} for approximate inference in the form of a sequential Bayesian filter, 
%described in Eq.~\ref{eq_pf}
 \begin{equation}
　　\label{eq_pf}
　　　Bel(q_t) \propto p(z_t|q_t)\sum_{j}p(q_t^{(j)}|q_{t-1}^{(j)})Bel(q_{t-1}^{(j)})
\end{equation}
where a collection of $n$ weighted particles $\{q_t^{(j)},w_t^{(j)}\}_{j=1}^{n}$ is used to represent the pose belief $q_t$. 

%We employ a generative sampling based local search method to infer the pose of the transparent object. This method is inspired by the sampling methods, such as boostrap filter~\cite{gordon1993novel}, which offers robustness and versatility over the search space.

Each particle ${q_t^{(j)}}$ is a hypothesized six-DoF pose of the object and is associated with the weight ${w_t^{(j)}}$ indicating how likely the sample is to be close to the actual pose. The initial samples are generated by uniformly sampling the six-DoF pose with identical weight. The weight of each sample is then calculated by using the observation likelihood function described in the next paragraph. With the computed weights, an importance sampling with resampling procedure is performed to concentrate hypothesized particles to more weighted range. For state transition, each particle will be perturbed by a zero-mean Gaussian distribution in the space of six-DoF in the action model.  This inference can be naturally extended to the case of tracking
%if enough computation power is allowed and 
with an explicit action model and observations over time. In our implementation, the process will iteratively repeat until the average weight is above a chosen threshold for taking an estimate. 

Our likelihood function measures the score of a sample's rendered depth image for a scene DLV. The z-buffer of a 3D graphics engine is used to render each sample into a depth image for comparison with the observation. This rendered depth image, represented as $z^{(j)}$, is mapping back to DLV to find the corresponding depth likelihood interval $[l_n,l_m)$. Here, we use an interval because the rendered depth value for a certain pixel may not exactly match its discretized depth value. After finding the corresponding interval, the depth likelihood is calculated using linear interpolation:
\begin{equation}
L(\mathbf{x}_\rho,l_n) = L(\mathbf{x}_\rho,l_n) + \frac{(l-l_n)(L(\mathbf{x}_\rho,l_m)-L(\mathbf{x}_\rho,l_n))}{l_m-l_n}
\end{equation}
For the rendered image, with every rendered pixel having non-zero (vaild) depth value $l_i$, the score for this depth image can be expressed as:
\begin{equation}
L(z_t) = \frac{\sum_i L(\mathbf{x}_\rho,l_i)}{N}
\end{equation}
where $N$ is the number of valid depths in the rendered image.

\section{Results}
\begin{figure}[t!]
   \centering
      \includegraphics[scale=0.76]{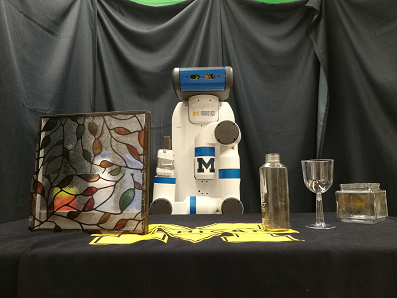}
      \caption{Test objects for evaluating PMCL 6D pose estimation include: (to the left)  opaque objects %(lint remover and bleach bottle) 
behind a partially opaque translucent surface (a stained glass window film), and (to the right) transparent objects. 
%(glass bottle, plastic wine glass, and glass jar).
}
      \label{experiment_set}
\end{figure}

\begin{figure}
\centering
\begin{floatrow}
\centering
\subfloat[]{
\includegraphics[width=0.45\columnwidth]{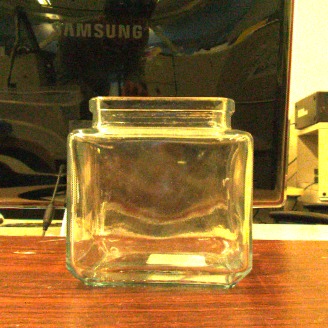}
\label{exp1}
}
\hspace{-0.5cm}
\subfloat[]{
\includegraphics[width=0.45\columnwidth]{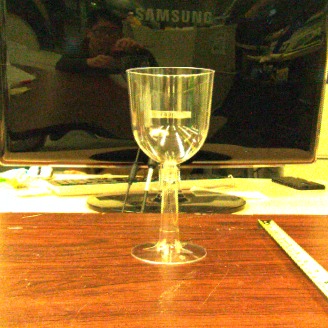}
\label{exp2}
}
\end{floatrow}

\begin{floatrow}
\subfloat[]{
\includegraphics[width=0.45\columnwidth]{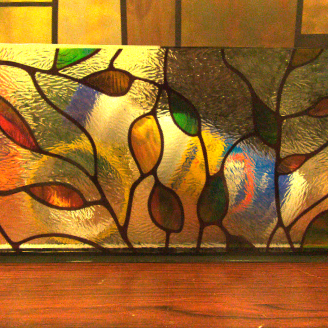}
\label{exp3}
}
\hspace{-0.5cm}
\subfloat[]{
\includegraphics[width=0.45\columnwidth]{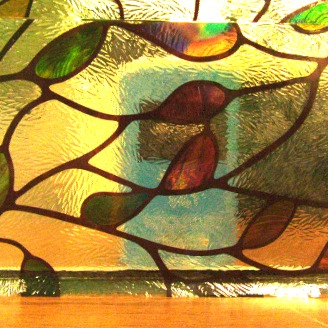}
\label{exp4}
}

\end{floatrow}
\caption{Two types of scene for localizing object poses. (a-b) the scene with a single transparent object with an opaque but possible reflective background objects. (c-d) the scene with opaque objects behind translucent non-transparent surfaces }
\label{fig:estimation_scenes}
\end{figure}

We now present results for our implementation of PMCL for object localization and grasping in environments with different forms of translucency.  We have implemented PMCL using observations from a Lytro light field camera mounted on the wrist of a Michigan Progress Fetch robot (Figure~\ref{experiment_set}).  These results consider pick-and-place grasping in two types of scenes with: 1) a single transparent object with an opaque but possible reflective background objects (Figures~\ref{exp1}, ~\ref{exp2}), and 2) opaque objects behind translucent non-transparent surfaces (Figures~\ref{exp3}, ~\ref{exp4}).%\ocjf{provide pointers here to your visual results for each of these types of scenes.}

Our implementation uses the Lytro on-chip wifi to trigger the shutter remotely and receive raw image data. We are currently unable to capture video with this triggering system. Calibration and sub-aperture images are generated using the methods described by Bok et al.~\cite{bok2017geometric}. This toolbox generates $9\times 9$ sub-aperture images, where the image at index $(5,5)$ is deemed the center view image. Each sub-aperture image has resolution $328\times 328$. During DLV construction, we disregard edge sub-aperture images due to strong color distortion and pixel shifting artifacts.%\ocjf{are we still using edge images in this case? I would assume we would have 25 images considered out of 49 total?}.
%we only use the 49 images indexed between $(2,2)$ to $(8,8)$ \ocjf{are we still using edge images in this case? I would assume we would have 25 images considered out of 49 total?} due to artifacts of color distortion and pixel shi edge images have strong color distortion and pixel shifting. 
% \begin{figure}[thpb]
%    \centering
%       \includegraphics[scale=0.25]{sub_aperture}
%       \caption{Left: raw light field image; Right: part of sub-apertures, the image with red rectangle is center view}
%       \label{raw_sub}
%    \end{figure}
   
Our PMCL algorithm is implemented on CUDA and OpenGL.  This implementation ran on a Ubuntu 14.04 operating system with a Titan X graphics card and CUDA 8.0. The light field camera calibration, sub-aperture images extraction, and DLV construction ran in MATLAB.  The chosen parameters for building the DLV were 
$\beta=0.5$, $\tau_1=0.5$, $\tau_2=0.5$, $l=75$, $N_{lm}=2$, and $K_{lm}=2$.  The Monte Carlo localization process ran on the GPU with 100 particle samples over 500 iterations.
%PMCL assumes the object's geometric model and 3D workspace are given {\it a priori}. The algorithm is implemented with CUDA-OpenGL interoperation to 
With an assumed object geometry, our implementation renders all the particle hypotheses on the GPU.  These renderings can be accessed by the CUDA kernels to compute the corresponding weights. %The majority of our computation is performed directly on the GPU, with little data transfer  between main CPU memory.  This efficiency enables the deployment of more particles for pose estimation.  
Our implementation additionally assumes a given 3D region of interest on the object pose in workspace.

For robot control, we use our custom manipulation pipeline developed by the Laboratory for Progress.  This pipeline uses our implementation of handle grasp localization as proposed by ten Pas and Platt~\cite{tenPas2015}. %\ocjf{can use the 2014 isrr citation instead?}.  
 This grasp localization returns an end-effector pose for grasping from an estimated object pose with a given geometric model.  Grasping is then executed for this end-effector pose using TRAC-IK~\cite{beeson2015trac} and MoveIt!~\cite{sucan2013moveit} for inverse kinematics and motion planning. 
%to execute the trajectory. Grasp localization is from an estimated The end effector's pose is calculated by grasp pose generation pipeline based on Darboux frame (surface normal and principal curvature axes) of each object geometry model proposed by \cite{ten2016localizing}.

%\section{Results}

%We examed our method on two types of the scene as shown in Figure \ref{experiment_set}: 1) scenes with single transparent object; 2) scenes with opaque objects behind the partial opaque surfaces. Our system ran on a Ubuntu 14.04 system with Titan X Graphics card and CUDA 8.0. The light field camera calibration, sub-apertures image extraction, and DLV construction ran on MATLAB. The parameter for building DLV is shown in Tab.\ref{DLV_para}.

\ignore{
\begin{table}[h]
\caption{DLV Parameters}
\label{DLV_para}
\begin{center}
\begin{tabular}{|c||c||c||c||c||c|}
\hline
$\beta$ &$\tau_1$ &$\tau_2$ &$l$ &$N_{lm}$ & $K_{lm}$  \\
\hline
$0.5$ &$0.5$ &$0.5$ &$75$ &$2$ & $2$  \\
\hline
% \hline
% $\beta$ & Two\\
% \hline
% $\gamma$ & Four\\
% \hline
% \hline
% $\tau_1$ & Two\\
% \hline
% $\tau_2$ & Four\\
% \hline
% $N_{lm}$ & Two\\
% \hline
% $K_{lm}$ & Four\\
% \hline
\end{tabular}
\end{center}
\end{table}
}

%The Monte Carlo localization process ran on GPU with 100 samples for 500 iterations. %For every 100 iterations,  
%we will state a new round and re-initialize the particle. For the last round we will re-initialize the particles use the best result among previous 4 rounds.

To evaluate the pose estimation accuracy of our algorithm, we used two methods to collect ground-truth object poses.  For objects behind the window covered by stained glass film, we captured point clouds by removing the glass and using Asus Xtion Pro RGB-D on the robot. Object models were then fit manually to determine ground truth pose values. For transparent objects, their surfaces were covered with opaque tape to generate point clouds for ground truth annotation.

\begin{figure}
\centering
\begin{floatrow}
\centering
\subfloat[]{
\includegraphics[width=0.52\columnwidth]{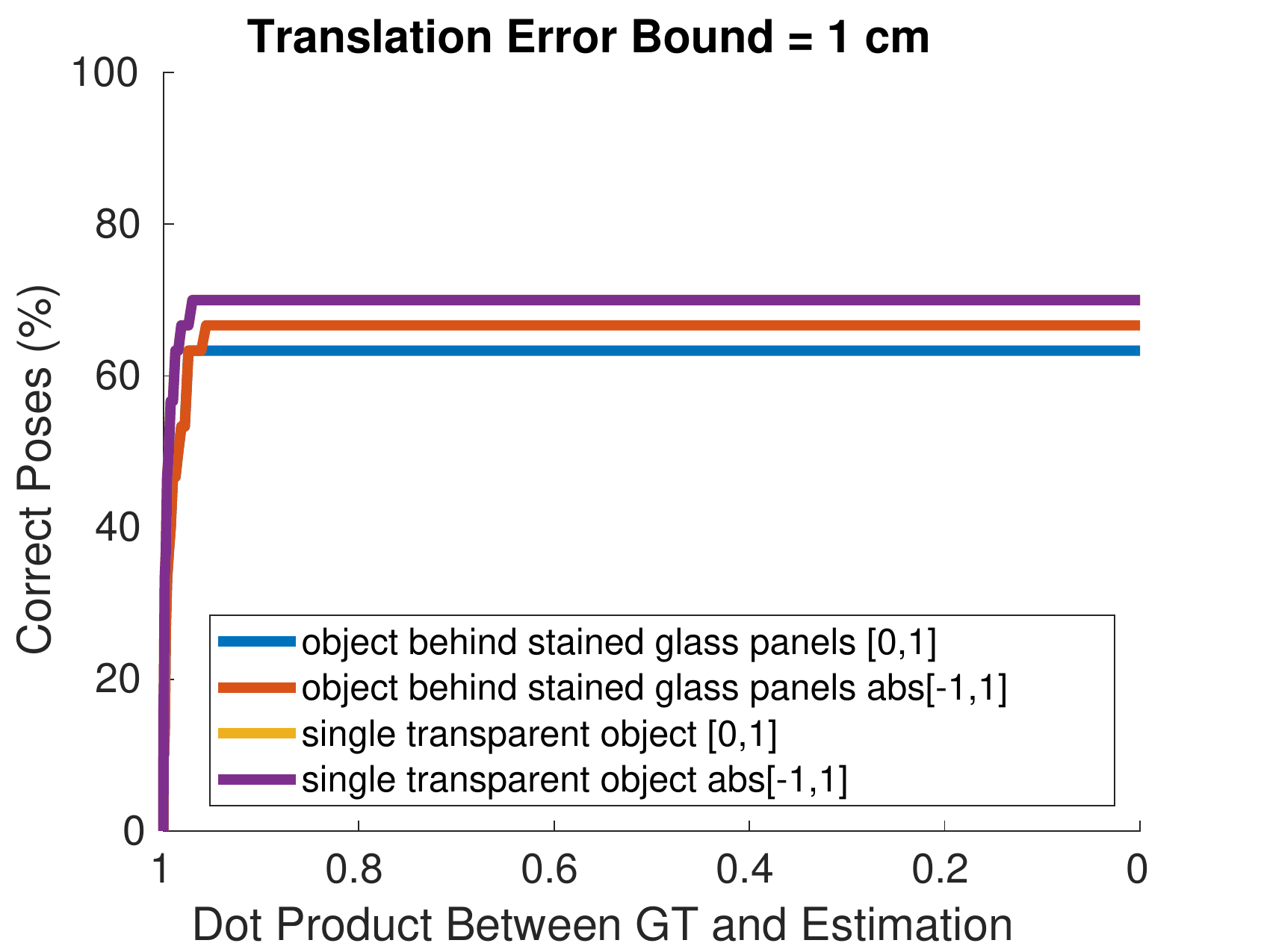}
\label{pose_err_1cm}
}
\hspace{-1cm}
\subfloat[]{
\includegraphics[width=0.52\columnwidth]{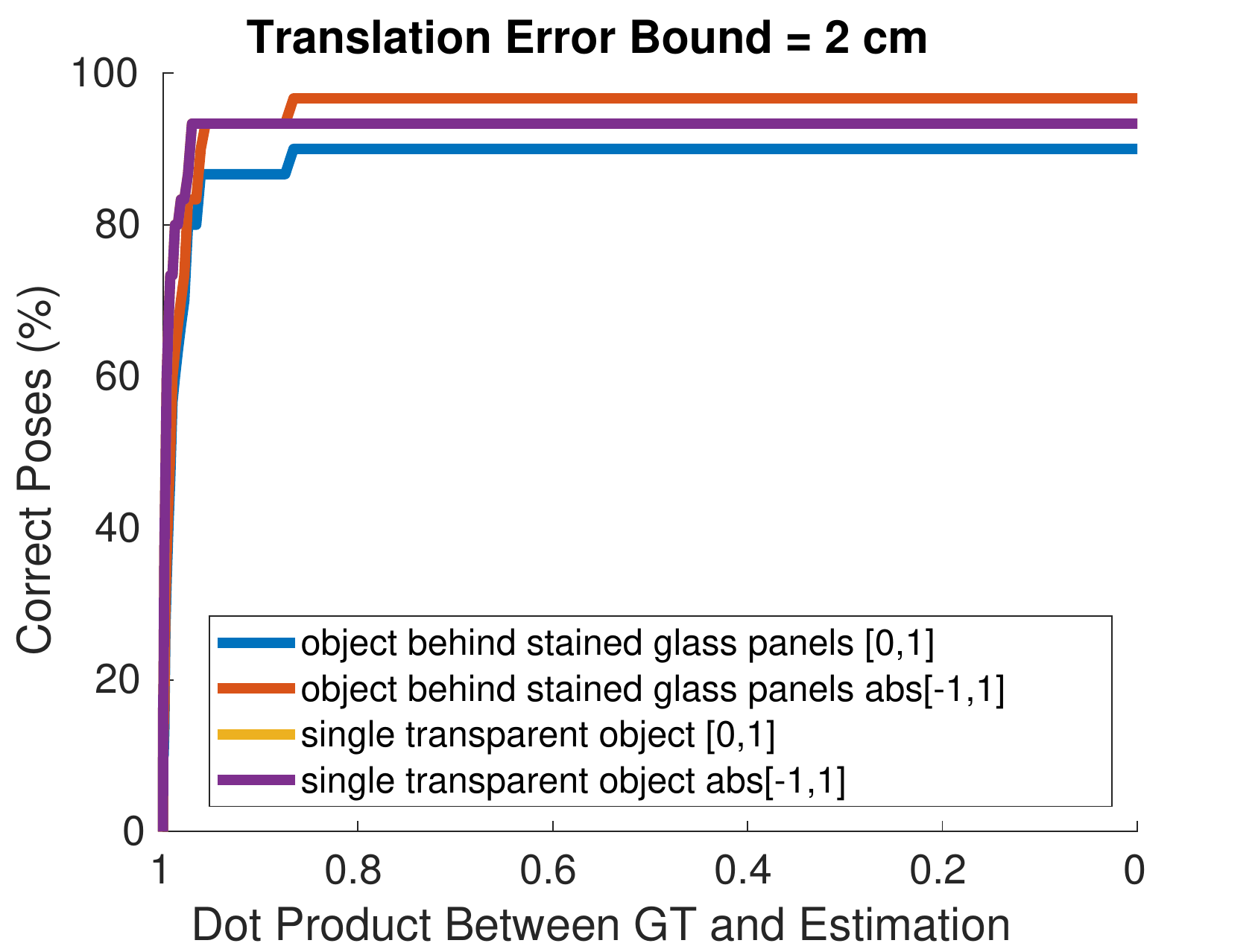}
\label{pose_err_2cm}
}
\end{floatrow}

% \begin{floatrow}
% \subfloat[]{
% \includegraphics[width=0.52\columnwidth]{figures/metric05.eps}
% \label{pose_err_5cm}
% }
% \hspace{-1cm}
% \subfloat[]{
% \includegraphics[width=0.52\columnwidth]{figures/metric10.eps}
% \label{pose_err_10cm}
% }

%\end{floatrow}
\caption{%\ocj{these plots are still confusing and somewhat annoying. can we pick one of these to show, or at least a reasonable synthesis of these into a single figure?} 
The percentage of correctly localized object under different thresholds for the object behind a stained glass panel and a single transparent object. In each plot, the translation error bound is fixed to 1cm (a) and 2cm (b). The x-axis is the decreasing dot product bound indicate the error between ground truth and estimated result. The y-axis is the percentage of correctly localized objects. For each type of scene, these plots consider two types of rotation error ranges: $[0,1]$ in dot product space indicates for [90,0] in degrees, and the absolute value [-1,1] in dot product space indicates [180, 0] in degrees.}
% Each point in the curve indicates the accuracy with a fixed translation and rotation error bound.}
\label{fig:progress_exp_pose_err}
\end{figure}

%\subsection{\ocj{Pose Estimation of Transparent Objects}}
%\subsection{\ocj{Pose Estimation under Diffuse Translucent Occlusion}}
%\ocj{will you break these results down by the two types of translucency?}
\subsection{Pose Estimation Results}
We evaluate our proposed algorithm on six scenes and run ten trials for each. Two types of error are applied to evaluate our pose estimation accuracy:
\begin{itemize}
\item Translation error: defined as the Euclidean distance between estimated object position $(x,y,z)$ and ground truth position $(x_{gt},y_{gt},z_{gt})$

% \item Rotation error: defined as Euler Euclidean distance between estimated object orientation $(roll,pitch,yaw)$ and ground truth orientation $(roll_{gt},pitch_{gt},yaw_{gt})$.
 \item Rotation error: defined as dot product between ground truth pose z-axis and estimated pose z-axis.
We assume the objects are rotational symmetric along z-axis.
\end{itemize}
We consider an object is correctly localized when both translation and rotation errors fall into a certain threshold. Figure  \ref{fig:progress_exp_pose_err} establishes our estimation accuracy on two types of the scene. 
% Each sub-plot shows the correct localization percentage respect to a changing rotation error threshold and a fixed translation error threshold (1cm and 2cm). 
% The Table~\ref{pose_acc} shows the overall estimation result for two types of the scene.
% Plot and table shows the satisfactory result of our method. 
For the single transparent object, the all rotation error in dot product space laid in [0,1], which leads to the overlapping of yellow and purple lines in both plots.
For an object behind stained glass panels, the estimated poses sometimes have 180 degree flipping, a negligible form of error assuming symmetry.
%which leads to different between blue and orange lines in both plots. We consider both correct pose and 180 flipping pose as good manipulation pose according to our experiment result.
% because we found both situations yield successful grasping in manipulation experiment.

% \begin{table}[h]
% \caption{Pose Estimation Accuracy}
% \label{pose_acc}
% \centering
% \begin{tabular}{|c|c|c||c|c|}
% \hline
% Scene & \multicolumn{2}{c||}{Trans Error} & \multicolumn{2}{c|}{Rot Error} \\
% & \multicolumn{2}{c||}{(Meter)} & \multicolumn{2}{c|}{(Degree)} \\
% \hline
% & Mean & Std & Mean & Std \\ 
% \hline
% Transparent Objects & 0.0094 & 0.007 & 4.902 & 4.07 \\ 
% \hline
% Translucent Surfaces & 0.0071 & 0.008 & 19.15 & 41.56 \\ 
% \hline
% \end{tabular}
% \end{table}

%Figure \ref{pose_err_2cm} shows that with 2cm translation error, our method achieve more than 80\% localization accuracy for both types of the scene. With low translation error threshold (Figure \ref{pose_err_1cm}), our method performs better on transparent object than the object behind the glass because the window film attached on the glass will blur the edge of the object behind which will make the object looks bigger and closer than the ground truth.
\subsection{Manipulation Results}
\begin{figure*}[!t]
\includegraphics[width=0.98\columnwidth]{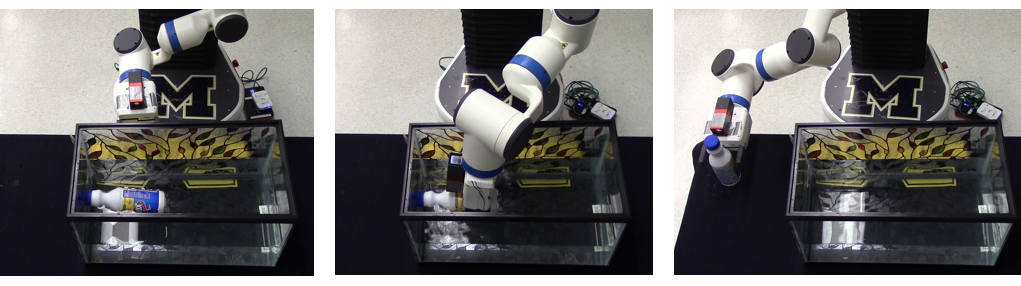}
\caption{ The robot executes pick-and-place action for the bleach bottle floating on the water. The bleach bottle is inside the aquatic tank so it is occluded by the stained glass from the camera view.}
\label{exp:acquatic}
\end{figure*}

We succeed in demonstrating our method in two challenge scenarios for manipulation\footnote{Video available on \url{https://youtu.be/Fu\_SVRXsdU8}},
\begin{enumerate}
\item Pick-and-place glass cup from a sink with running water
\item Pick-and-place bleach bottle from an aquatic tank covered with private window film. 
\end{enumerate}
The scenarios are shown in Figure~\ref{exp:sink} and Figure~\ref{exp:acquatic}. We attach the Lytro camera to the wrist of the robot and add extra link for it. For both scenarios, the robot moves its arm to the appropriate area to capture the light field images, from which the DLV is calculated. Our PMCL then performs estimation to infer the pose of the object and the final estimation is taken to transform the pre-calculated grasp poses in robot base link. With the accurate pose estimation, the robot is able to pick up objects from both aquatic tank and sink and place the objects on the desired location.

\section{Conclusion}
In this paper, we present Plenoptic Monte Carlo Localization for localizing object pose in the presence of translucency from plenoptic (light-field) observations. We propose a new depth descriptor, the Depth Likelihood Volume, to address the uncertainties from the translucency by generating possible depth likelihoods for each pixel. We show that by using the Depth Likelihood Volume within a Monte Carlo object localization algorithm our method is able to accurately localize objects with translucent materials and objects occluded by layered translucency and perform manipulation.

\addtolength{\textheight}{-12cm}   % This command serves to balance the column lengths

%%%%%%%%%%%%%%%%%%%%%%%%%%%%%%%%%%%%%%%%%%%%%%%%%%%%%%%%%%%%%%%%%%%%%%%%%%%%%%%%

%%%%%%%%%%%%%%%%%%%%%%%%%%%%%%%%%%%%%%%%%%%%%%%%%%%%%%%%%%%%%%%%%%%%%%%%%%%%%%%%

\bibliographystyle{abbrv}
\bibliography{big}

\end{document}